\documentclass[conference]{IEEEtran}
\IEEEoverridecommandlockouts

\usepackage{cite}
\usepackage{units}
\usepackage{tabularx} 

\usepackage[%
colorlinks = false,
linkcolor = blue,
urlcolor  = blue,
citecolor = blue,
anchorcolor = blue]{hyperref}

\usepackage[export]{adjustbox}

\usepackage{siunitx}
\usepackage[nolist]{acronym}

\usepackage{xcolor}

\usepackage{tikz}
\pgfdeclarelayer{foreground}
\pgfsetlayers{background, main, foreground}
\usetikzlibrary{quotes, angles, backgrounds, arrows, automata, shapes, positioning, calc, through, 
spy, decorations.markings}

\usepackage{aircraftshapes} 

\tikzset{
	imglabel/.style={
		rectangle,
		inner sep=2pt,
		text=black,
		minimum height=1em,
		text centered,
		fill=white,
		fill opacity=1.0,
		text opacity=1,
		anchor=south west,
	},
}

\tikzset{
	state/.style={
		rectangle,
		draw=black, very thick,
		minimum height=1.0em,
		text centered,
	},
}

\usepackage{textcomp}
\usepackage{graphicx} 
\usepackage{epsfig} 
\usepackage{pgfplots}
\usepackage{caption}
\usepackage{subcaption}

\def\BibTeX{{\rm B\kern-.05em{\sc i\kern-.025em b}\kern-.08em
		T\kern-.1667em\lower.7ex\hbox{E}\kern-.125emX}}

\usepackage{amsmath, amssymb, amsfonts, mathtools}

\newcommand{\splitatcommas}[1]{%
	\begingroup
	\ifnum\mathcode`,="8000
	\else
	\begingroup\lccode`~=`, \lowercase{\endgroup
		\edef~{\mathchar\the\mathcode`, \penalty0 \noexpand\hspace{0pt plus 1em}}%
	}\mathcode`,="8000
	\fi
	#1%
	\endgroup
}

\newcommand{\Lim}[1]{\raisebox{0.5ex}{\scalebox{0.8}{$\displaystyle \lim_{#1}\;$}}}

\DeclareMathOperator*{\minimize}{minimize}
\DeclareMathOperator*{\maximize}{maximize}

\def\xx{\mathsf{x}}
\def\yy{\mathsf{y}}
\def\zz{\mathsf{z}}

\newcommand\copyrighttext{%
	\small \begin{center} \color{red} \textcopyright\,2021 IEEE. Personal use of this material is 
	permitted. Permission from IEEE must be obtained for all other uses, in any current or future 
	media, including reprinting/republishing this material for advertising or promotional 
	purposes, creating new collective works, for resale or redistribution to servers or lists, or 
	reuse of any copyrighted component of this work in other works. \end{center}}
\newcommand\copyrightnotice{%
	\begin{tikzpicture}[remember picture,overlay]
	\node[anchor=south,yshift=25.6cm] at (current page.south) 
	{\color{white}\fbox{\parbox{\dimexpr\textwidth-\fboxsep-\fboxrule\relax}{\copyrighttext}}};
	\end{tikzpicture}%
}

\title{\copyrightnotice \LARGE \bf
	Optimum Trajectory Planning for Multi-Rotor UAV Relays with Tilt and Antenna Orientation 
Variations
}

\author{Daniel Bonilla Licea$^1$, Giuseppe Silano$^1$, Mounir Ghogho$^2$, and Martin 
	Saska$^1$
	\thanks{$^1$Daniel Bonilla Licea, Giuseppe Silano, and Martin Saska are with the Faculty of 
	Electrical Engineering, Czech Technical University in Prague, Czech Republic (email: {\tt\small 
			\{bonildan, giuseppe.silano, martin.saska\}@fel.cvut.cz}).}
	\thanks{$^2$Mounir Ghogho is with the International University of Rabat, Morocco (email: 
	{\tt\small mounir.ghogho@uir.ac.ma}).}
	\thanks{This work was partially funded by the European Union's Horizon 2020 research and 
	innovation programme AERIAL-CORE under grant agreement no. 871479, by CTU grant no. 
	SGS20/174/OHK3/3T/13, and by the Czech Science Foundation (GAČR), within research project no. 
	20-10280S.}
}    

\begin{document}
	
\maketitle
\thispagestyle{empty}
\pagestyle{empty}



\begin{acronym}
	\acro{AoA}[AoA]{Angle of Arrival}
	\acro{AoD}[AoD]{Angle of Departure}
	\acro{AWGN}[AWGN]{Additive White Gaussian Noise}
	\acro{BS}[BS]{Base Station}
	\acro{FDD}[FDD]{Frequency-Division Duplexing}
	\acro{LoS}[LoS]{Line of Sight}
	\acro{SIL}[SIL]{Software-in-the-loop}
	\acro{SNR}[SNR]{Signal-to-Noise Ratio}
	\acro{UAV}[UAV]{Unmanned Aerial Vehicle}
\end{acronym}



\begin{abstract}
	
	Multi-rotor~\acp{UAV} need to tilt in order to move; this modifies the~\ac{UAV}'s antenna 
	orientation. We consider the scenario where a multi-rotor~\ac{UAV} serves as a communication 
	relay between a~\ac{BS} and another~\ac{UAV}. We propose a framework to generate feasible 
	trajectories for the multi-rotor~\ac{UAV} relay while considering its motion dynamics and the 
	motion-induced changes of the antenna orientation. The~\ac{UAV} relay's trajectory is optimized 
	to maximize the end-to-end number of bits transmitted. Numerical simulations in MATLAB and 
	Gazebo show the benefits of accounting for the  antenna orientation variations due to 
	the~\ac{UAV} tilt. 
	
	
\end{abstract}






\vspace{-1em}
\section{Introduction}
\label{sec:intro}

Communications-aware robotics is gaining momentum as evidenced by the steady increase in 
publications by the robotics~\cite{Gasparri2016TRO, Varadharajan2020RAL,Licea2017TRO,Licea2020TRO} 
and the communications~\cite{Zeng2017TWC, Wu2018TWC, Dabiri2020CL} communities dealing with mobile 
robots and communications issues. One reason behind this growing interest is the emergence of the 
5G technology that aims to integrate~\acfp{UAV} in the cellular communications 
network~\cite{Mishra2020CN, Zeng2019PIEEE}. 

It is in this context that we encounter the problem of robotic relay trajectory planning where 
mobile robots, acting as communications relays, have their trajectories optimized according to some 
communications criteria. One popular type of robots used as relays is the multi-rotor~\ac{UAV} 
which requires to tilt in order to move~\cite{Mahony2012RAM}. Thus, when two multi-rotor~\acp{UAV}, 
equipped with a single fixed antenna each, communicate while moving, the resulting~\ac{SNR} will 
depend not only on their positions but also on their tilt. As the tilt of the receiver~\ac{UAV} 
changes, the~\ac{AoA}, measured w.r.t. the antenna reference frame, also changes and so does the 
antenna gain experienced by the received signal. In other words, in the case of 
multi-rotor~\acp{UAV} communications, the quality of the channel depends not only on the~\ac{UAV}'s 
position but also on its orientation (called attitude in the robotics literature).

In some cases, multi-rotor~\acp{UAV} hover while relaying data; the~\ac{UAV}'s orientation remains 
constant, and so there are no changes in the communications channel induced by the~\ac{UAV}'s 
orientation. For instance, in~\cite{LiceaICASSP2020}, we studied the scenario where a quad-rotor 
hovers to collect data from ground sensors; although the~\ac{UAV} antenna's radiation pattern was 
taken into account, the antenna orientation remained constant. 

In other applications the multi-rotor~\acp{UAV} relay communicates on the move. The existing 
studies however disregard the effect of the~\ac{UAV} tilt on the antenna orientation. 
In~\cite{Wu2018TWC}, the authors consider a multi~\ac{UAV} communications system, but the~\ac{UAV} 
dynamics are oversimplified, and the antenna radiation pattern is disregarded. 
In~\cite{Ahmed2020CL}, the authors consider a~\ac{UAV} relay between ground users, but the effect 
of the antenna radiation pattern is overlooked. In~\cite{Licea2019CDC,Licea2020TRO}, we considered 
scenarios where the multi-rotor~\ac{UAV} communicates while moving; we considered more realistic  
dynamic models for the multi-rotor~\acp{UAV}, but we disregarded the antenna radiation pattern.

Works similar to ours involving multi-rotor~\acp{UAV} relays are found 
in~\cite{ZhangAccess2018,LiuICCC2019}. However, these papers do not consider the effect of the 
antenna radiation pattern and use oversimplified dynamic models for the~\acp{UAV}. 
In~\cite{Zhou2020ACM}, the authors consider the problem of trajectory planning for a ground robotic 
relay in an indoor scenario, and do take into consideration the antenna radiation pattern of a 
mobile end-target. 

In this paper, we optimize the trajectory of a multi-rotor~\ac{UAV} relay between a ground~\acf{BS} 
and another moving multi-rotor~\ac{UAV}. We consider the antennas' orientation changes, due to 
the~\acp{UAV} changing tilt, on the communications channel. This is done by simultaneously 
considering the~\ac{UAV} dynamical model and its antenna radiation pattern. To the authors' best 
knowledge, this is the first time that, in the context of multi-rotor of~\acp{UAV} communications, 
the~\ac{UAV} tilt-induced changes in the antenna orientation is accounted for in the trajectory 
design.



\subsection{Organization}
\label{sec:organization}

In Sec.~\ref{sec:motionPrimitives}, we describe the quad-rotor's dynamics and the communications 
system. Sec.~\ref{sec:ProbStat} presents and analyses the communications-aware trajectory planning 
problem to be solved. Simulations results are presented in Sec.~\ref{sec:Sim}. Finally, conclusions 
are drawn in Sec.~\ref{sec:Conc}.



\subsection{Notation}
\label{sec:notation}

$c_\bullet$ and $s_\bullet$ are short notations for  $\cos(\bullet)$ and $\sin(\bullet)$, 
$\prescript{k}{}{x}$ is the value of function $x$ at discrete time $k$, 
$\prescript{k}{}{\mathbf{p}}^{(n)}$ is the $n$ entry of vector $\mathbf{p}$ evaluated at time 
instant $k$.



\section{Preliminaries}
\label{sec:preliminaries}



\subsection{Quad-rotor dynamics}
\label{sec:motionPrimitives}

As mentioned above, in this work,~\acp{UAV} are taken to be multi-rotors, and especially 
quad-rotors. These~\acp{UAV} have the particularity that they can either stay still in the air by 
hovering, or move towards any desired destination, as long as their dynamics constraints are not 
violated. In order to move, the quad-rotor needs to tilt; its direction of movement ($\mathbf{p}$), 
velocity ($\mathbf{v}$) and acceleration ($\mathbf{a}$) depend on its Euler 
angles~\cite{Siciliano2016Handbook}: roll ($\varphi$), pitch ($\vartheta$) and yaw ($\psi$).

We consider a discrete-time dynamic model for the quad-rotor; let $T_s \in \mathbb{R}_{\geq 0}$ and 
$T \in \mathbb{R}_{\geq 0}$ denote the~\ac{UAV} sampling period and trajectory time, respectively, 
and let $\mathbf{t}=[0, T_s, \dots, NT_s]^\top \in \mathbb{R}^{N+1}$, with 
$\prescript{k}{}{\mathbf{t}} = kT_s$, $k \in \mathbb{N}_{\geq 0}$, and $NT_s=T$. We also define the 
state $\mathbf{x}$ and control $\mathbf{u}$ sequences as $\prescript{k}{}{\mathbf{x}} = 
[\prescript{k}{}{\mathbf{p}}^{(1)}, \prescript{k}{}{\mathbf{v}}^{(1)}, 
\prescript{k}{}{\mathbf{p}}^{(2)}, \prescript{k}{}{\mathbf{v}}^{(2)}, 
\prescript{k}{}{\mathbf{p}}^{(3)}, \prescript{k}{}{\mathbf{v}}^{(3)}]^\top$ and 
$\prescript{k}{}{\mathbf{u}} = [\prescript{k}{}{\mathbf{a}}^{(1)}, 
\prescript{k}{}{\mathbf{a}}^{(2)}, \prescript{k}{}{\mathbf{a}}^{(3)}]^\top$, where 
$\prescript{k}{}{\mathbf{p}}^{(j)}$, $\prescript{k}{}{\mathbf{v}}^{(j)}$, and 
$\prescript{k}{}{\mathbf{a}}^{(j)}$, with $j=\{1, 2, 3\}$, represent the vehicle's position, 
velocity, and acceleration at time instant $k$ along the $j$-axis of the inertial frame $O_W$, 
respectively.

In~\cite{Mueller2015TRO}, the authors present motion primitives to design trajectories that satisfy 
the~\ac{UAV}'s dynamic constraints. This method allows to generate feasible quad-rotor motion 
primitives. This method provide the following splines that we will use to account for the dynamics 
of the quad-rotor~\ac{UAV}:
\begin{equation}\label{eq:splines}
	\resizebox{1\hsize}{!}{$
		\begin{bmatrix}
		\prescript{k+1}{}{\mathbf{p}}^{(j)} \\
		\prescript{k+1}{}{\mathbf{v}}^{(j)} \\
		\prescript{k+1}{}{\mathbf{a}}^{(j)} 
		\end{bmatrix}
		= 
		\begin{bmatrix}
		\frac{\alpha}{120} \prescript{k}{}{\mathbf{t}}^5 + \frac{\beta}{24} 
		\prescript{k}{}{\mathbf{t}}^4 + \frac{\gamma}{6}
		\prescript{k}{}{\mathbf{t}}^3 + \prescript{k}{}{\mathbf{a}}^{(j)} 
		\prescript{k}{}{\mathbf{t}}^2 + \prescript{k}{}{\mathbf{v}}^{(j)} 
		\prescript{k}{}{\mathbf{t}} + \prescript{k}{}{\mathbf{p}}^{(j)} \\
		\frac{\alpha}{24} \prescript{k}{}{\mathbf{t}}^4 + \frac{\beta}{6} 
		\prescript{k}{}{\mathbf{t}}^3 + \frac{\gamma}{2} 
		\prescript{k}{}{\mathbf{t}}^2 + \prescript{k}{}{\mathbf{a}}^{(j)} 
		\prescript{k}{}{\mathbf{t}} + \prescript{k}{}{\mathbf{v}}^{(j)} \\
		\frac{\alpha}{6} \prescript{k}{}{\mathbf{t}}^3 + \frac{\beta}{2} 
		\prescript{k}{}{\mathbf{t}}^2 + \gamma \, \prescript{k}{}{\mathbf{t}} + 
		\prescript{k}{}{\mathbf{a}}^{(j)}
		\end{bmatrix},
		$}
\end{equation}
where $\alpha$, $\beta$ and $\gamma$ are design parameters that determine the behaviour at the 
start and end points~\cite[Appx.~A]{Mueller2015TRO}.



\subsection{Communications System}
\label{sec:systemModel:Comms}

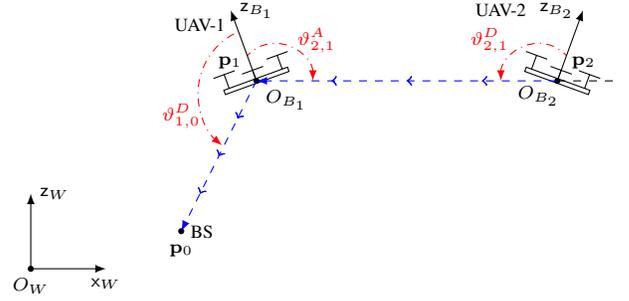
\begin{figure}
	\begin{center}
		\scalebox{1}{
			\begin{tikzpicture}
			\tikzset{->-/.style={decoration={
						markings,
						mark=at position #1 with {\arrow{>}}}, postaction={decorate}}
			}
			
			\node (p_UAV1) at (-0.35,0.25) {\scriptsize $\mathbf{p}_1$};
			\node (p_UAV2) at (4.35,0.25) {\scriptsize $\mathbf{p}_2$};
			\node (p_BS) at (-1,-2.25) {\scriptsize $\mathbf{p}_0$};
			
			\node (UAV1) [quadcopter side, fill=white, draw=black, minimum width=1.5cm, rotate=20] 
			at (0,0) {};
			\node at (-0.75,0.75) [text centered]{\scriptsize UAV-1};
			
			\node (UAV2) [quadcopter side, fill=white, draw=black, minimum width=1.5cm, rotate=-20] 
			at (4,0) {}; 
			\node at (3.25,0.95) [text centered]{\scriptsize UAV-2};
			
			\draw[red, dash dot, latex-] (0.75, 0) arc (0:138:0.5cm) node at (0.80, 0.25) 
			[above]{\scriptsize $\vartheta^A_{2,1}$};
			
			\draw[red, dash dot, -latex] (-0.3, 0.625) arc (118:230:0.90cm) node at (-1, -0.75) 
			[above]{\scriptsize $\vartheta^D_{1,0}$};
			
			\draw[red, dash dot, latex-] (3.25, 0) arc (180:42:0.5cm) node at (3.10, 0.25) 
			[above]{\scriptsize $\vartheta^D_{2,1}$};
			\draw[dashed] (4,0) -- (4.75,0);
			
			\node (circle) at (UAV1) [circle, draw, scale=0.2, fill=black] {};
			\node (circle) at (UAV2) [circle, draw, scale=0.2, fill=black] {};
			\node (circle) at (-1,-2) [circle, draw, scale=0.2, fill=black] {};
			\draw (-1,-2) node[right]{\scriptsize $\text{BS}$};
			\draw[-latex, dashed, blue, ->-=.5, ->-=.25, ->-=.75] (UAV2) -- (UAV1);
			
			\draw[dashed, blue, ->-=.5, ->-=.75, ->-=.25, -latex] (UAV1) -- (-1,-2); 
			
			\draw[-latex] (UAV2) node at (3.75,0.05) [below]{\scriptsize $O_{B_2}$} -- ($ (UAV2) + 
			(0.342020143, 0.939692621)$) node[left]{\scriptsize $\zz_{B_2}$};
			\draw[-latex] (UAV1) node[below right]{\scriptsize $O_{B_1}$} -- ($ (UAV1) + 
			(-0.342020143, 0.939692621)$) node[right]{\scriptsize $\zz_{B_1}$};
			
			\draw[-latex] (-3,-2.5) node[below]{\scriptsize $O_W$} -- (-3,-1.5) 
			node[right]{\scriptsize $\zz_W$};
			\draw[-latex] (-3,-2.5) -- (-2,-2.5) node[below]{\scriptsize $\xx_W$};
			\node (circle) at (-3,-2.5) [circle, draw, scale=0.2, fill=black] {};
			
			\end{tikzpicture}
		}
	\end{center}
	\vspace{-1em}
	\caption{Multi-rotor~\ac{UAV}-$1$ as a communications relay between the 
	multi-rotor~\ac{UAV}-$2$ and the~\ac{BS}.}
	\label{fig:scenario}
\end{figure}

The communications system consists of two communications links:~\ac{UAV}-$2\rightarrow$\ac{UAV}-$1$ 
and~\ac{UAV}-$1\rightarrow$\ac{BS}. Both~\acp{UAV} are equipped with a single antenna; we 
arbitrarily choose the half-wave dipole\footnote{Half-wave dipole is a common type of antenna.}, 
but the proposed method apply to other types of antennas. The $q$-th~\ac{UAV} antenna is located at 
its center of mass ${\mathbf{p}}_q$, and is aligned to its $\zz_{B_q}$ axis, see 
Fig.~\ref{fig:scenario}. The gain experienced by the wave transmitted by the $q$-th~\ac{UAV}'s 
is~\cite{Stutzman1981Book}:
\begin{equation}\label{CS:1}
	G_q(\vartheta)=\frac{ D \cos( \cos(\vartheta) \nicefrac{\pi}{2} ) }{\sin(\vartheta)}, \quad 
	q=\{1,2\},
\end{equation}
where $D$ is the half-wave dipole's directivity ($\approx 1.64$), $\vartheta$ is the~\ac{AoD} of 
the radiated wave measured w.r.t. the antenna's axis $\zz_{B_q}$. Note that~\eqref{CS:1} also 
describes the gain experienced by the received wave, in which case $\vartheta$ becomes 
the~\acf{AoA}.

To highlight the effect of the coupling between the~\acp{UAV}' tilt and its antenna orientation, we 
perform the following simplifications: we assume~\ac{LoS} for both communications links, we neglect 
small-scale fading, and we assume that the~\ac{BS} tracks~\ac{UAV}-$1$ using beamforming. Then, we 
model the communications channels by using the free space model and including the effect of the 
antennas' radiation pattern. Thus, the~\ac{UAV}-$2\rightarrow$\ac{UAV}-$1$ channel is modeled as: 
\begin{equation}
	\label{CS:2}
	r_1 = \left( \frac{G_2(\vartheta_{2,1}^{D}) G_1(\vartheta_{2,1}^{A})} {\lVert \mathbf{p}_2 - 
	\mathbf{p}_1 \rVert} \right ) s_2 + n_1,
\end{equation}
where $r_1$ and $s_2$ are the signals received and transmitted by the~\ac{UAV}-$1$ 
and~\ac{UAV}-$2$, respectively; $n_1$ is the zero-mean complex~\ac{AWGN} with power $\sigma_1^2$ 
generated at the~\ac{UAV}-$1$'s receiver; $\vartheta_{2,1}^{D}$ and $\vartheta_{2,1}^{A}$ are 
the~\ac{AoD} and~\ac{AoA} measured w.r.t. the axes  ${\zz}_{B_2}$ and ${\zz}_{B_1}$, respectively. 
Using simple geometry, we have that:
\begin{eqnarray}
	\vartheta_{2,1}^{D}&=&\arctan\left(\frac{h_{2,1}^{(3)}}{\sqrt{(h_{2,1}^{(1)})^2+(h_{2,1}^{(2)})^2}}\right)-\frac{\pi}{2}
	 \label{CS:4},\\
	\vartheta_{2,1}^{A}&=&\arctan\left(\frac{h_{1,2}^{(3)}}{\sqrt{(h_{1,2}^{(1)})^2+(h_{1,2}^{(2)})^2}}\right)-\frac{\pi}{2}
	 \label{CS:5},
\end{eqnarray}
where $h_{w,q}^{(r)}$ is the $r$-element of the vector $\prescript{W}{}{\mathbf{R}}_{B_q}( 
\mathbf{p}_w-\mathbf{p}_q)$ with $\prescript{W}{}{\mathbf{R}}_{B_q}$ the rotation matrix from the 
global ($O_W$) to the $q$-th~\ac{UAV} coordinate system~\cite{Siciliano2016Handbook}. The 
UAV-$1\rightarrow$\ac{BS} channel is obtained by performing the following changes on 
the~\ac{UAV}-$2\rightarrow$\ac{UAV}-$1$ channel equations: (i) exchanging the subindexes as follows 
$2\rightarrow 1$ and $1\rightarrow 0$; (ii) setting $G_0(\vartheta_{1,0}^{A}) = D_B$ for all 
$\vartheta_{1,0}^{A}$ (to model the beamforming implemented by the~\ac{BS}) where $D_B$ is the 
directivity of the main beam tracking the~\ac{UAV}-$1$. Finally, we denote $\xi_{2,1}$ and 
$\xi_{1,0}$ the~\acp{SNR} of the~\ac{UAV}-$2\rightarrow$\ac{UAV}-$1$ 
and~\ac{UAV}-$1\rightarrow$\ac{BS} channels, respectively.



\vspace{-1em}
\section{Problem statement \& Solution}
\label{sec:ProbStat}
\vspace{-1em}

While~\ac{UAV}-$2$ follows a trajectory $\mathcal{T}_2$ it must transmit data to the~\ac{BS}. To 
improve the communications and extend \ac{UAV}-$2$'s range of action, another~\ac{UAV} 
(i.e.,~\ac{UAV}-$1$) acting as a relay is integrated to the system. We assume  that~\ac{UAV}-$2$ 
communicates only with \ac{UAV}-$1$ which simultaneously relays\footnote{This can be achieved by 
using~\ac{FDD}.} the data to the~\ac{BS} located at $\mathbf{p}_0$. The end-to-end channel capacity 
of this system corresponds to the capacity of the channel having the poorest~\ac{SNR}. Now, given 
$\mathcal{T}_2$, we want to optimize the predetermined~\ac{UAV}-$1$ trajectory so as to maximize 
the number of bits transmitted from~\ac{UAV}-$2$ to the~\ac{BS} via~\ac{UAV}-$1$. This can be 
achieved by solving:
\begin{equation} 
	\label{eq:optimizationProblem}
	\begin{split}
		&\maximize_{\mathbf{p}_1, \mathbf{v}_1, \mathbf{a}_1}\ \ 
		\displaystyle\sum_{k=0}^N \min \Bigl({\log_2 \Bigl(1 + \prescript{k}{}{\xi}_{1,0} 
		\Bigr)}, {\log_2 \Bigl(1 + \prescript{k}{}{\xi}_{2,1} \Bigr)} \Bigr) \\
		&\quad \,\;\, \text{s.t.}~\quad\qquad \prescript{k}{}{\xi}_{1,0} = \frac{D_B^2 \, 
		G_1^2(\prescript{k}{}{\vartheta}_{1,0}^D) P}{\lVert \prescript{k}{}{\mathbf{p}}_1 - 
		\mathbf{p}_0 \rVert ^2\sigma^2_0}, \\
		&\qquad\qquad\qquad\, \prescript{k}{}{\xi}_{2,1} = \frac{G_1^2 
		(\prescript{k}{}{\vartheta}_{2,1}^A) \, G_2^2(\prescript{k}{}{\vartheta}_{2,1}^D)P} 
		{\lVert{ \prescript{k}{}{\mathbf{p}}_2 - \prescript{k}{}{\mathbf{p}}_1 \rVert^2 
		\, \sigma^2_1}}, \\
		&\qquad \,\;\, \qquad \,\;\,~\quad\; \lvert \prescript{k}{}{\mathbf{v}}^{(j)} \rvert \leq 
		\mathbf{v}^{(j)}_\mathrm{max}, \lvert \prescript{k}{}{\mathbf{a}}^{(j)} \vert  \leq 
		\mathbf{a}^{(j)}_\mathrm{max}, \\
		&\,\;\, \qquad \qquad \qquad\; \text{eq.}~\eqref{eq:splines}, \mathrm{for}\ j=\{1,2,3\},\\
		&\,\;\, \qquad \qquad \qquad\; \forall k=\{0,1, \dots,N-1\}, 
	\end{split}
\end{equation}
where $P$ is the power of the transmitted signal, $\min ({\log_2 (1 + 
\prescript{k}{}{\xi}_{1,0})}, {\log_2 (1 + \prescript{k}{}{\xi}_{2,1})} )$ is the 
normalized upper bound for the end-to-end data bit rate at discrete time $k$. The angles 
$\vartheta_{2,1}^A$ and $\vartheta_{2,1}^D$ are calculated with \eqref{CS:4}--\eqref{CS:5}, while 
$\vartheta_{1,0}^D$ is also calculated in a similar manner. We want the optimum trajectory (i.e., 
$\mathbf{p}_1^\star$, $\mathbf{v}_1^\star$, and $\mathbf{a}_1^\star$) obtained after solving the 
optimization problem~\eqref{eq:optimizationProblem} to be feasible. We ensure this with the last 
two lines of constraints in \eqref{eq:optimizationProblem} that take into account the dynamics 
of~\ac{UAV}-1 and its physical constraints, i.e., maximum velocity and acceleration. 

Now, we transform the maximization problem~\eqref{eq:optimizationProblem} into a more standard 
minimization problem\footnote{After many simulations and numerical analysis we observed that, in 
this particular case, the maximization problem was difficult to solve and produced erratic 
trajectories; the minimization formulation produces coherent trajectories. More analysis is 
required to understand this issue.}. We achieve this by first replacing the $\maximize$ with 
$\minimize$ action and then changing the cost function in~\eqref{eq:optimizationProblem} to:
\begin{equation} \label{eq:optimizationProblemRevised}
	\bar{J} = \sum_{k=0}^N\max\left(\frac{1}{\log_2 ( 1+\prescript{k}{}{\xi}_{1,0})}, 
	\frac{1}{\log_2(1 + \prescript{k}{}{\xi}_{2,1})} \right).    
\end{equation}
After transforming~\eqref{eq:optimizationProblem} into a minimization problem we note that the 
$\max$ function in~\eqref{eq:optimizationProblemRevised} makes the optimization problem NP-hard: to 
evaluate the $\max$ function in $\bar{J}$ for a single candidate trajectory, the optimization 
algorithm must compare $\prescript{k}{}\xi_{1,0}$ with $\prescript{k}{}\xi_{2,1}$ for 
all $k$. To solve this, we approximate the $\max$ function with a smooth function to ensure that 
the optimization problem is no longer NP-hard. We use the following approximation, see 
Appendix~\ref{Appendix:minApp}:
\begin{equation*} \label{eq:approximationCostFunction}
	\bar{J} \approx \sum_{k=0}^N\left(\frac{1}{\left(\log_2(1 + \prescript{k}{}{\xi}_{1,0}) 
	\right)^{p}} + \frac{1}{\left(\log_2( 1 + \prescript{k}{}{\xi}_{2,1}) \right)^{p}}\right)^{1/p}.
\end{equation*}
From~\eqref{CS:1}--\eqref{CS:4} we observe that $G_2^2(\vartheta_{2,1}^D)$ is a highly nonlinear 
function of $\mathbf{p}_1$; this adds local minima to the optimization problem. We alleviate this 
by approximating the antenna power gains as follows, see Appendix~\ref{Appendix:radApp}:
\begin{equation} \label{eq:prob.7}
	G_2^2(\vartheta_{2,1}^D) \approx D_d^2 \, \mathbf{g}(\vartheta_2, \varphi_2)^\top 
	\mathbf{v}_{2,1}, 
\end{equation}
where $\vartheta_2$ and $\varphi_2$ are the roll and pitch angles for~\ac{UAV}-$2$, respectively, 
$\splitatcommas{\mathbf{g}(\vartheta_2, \varphi_2) = [1, -s^2_{\vartheta_2}, -c^2_{\vartheta_2} 
s^2_{\varphi_2}, -c^2_{\varphi_2} c^2_{\vartheta_2}, 2c_{\vartheta_2} s_{\vartheta_2} 
s_{\varphi_2}, -2c_{\vartheta_2} c_{\varphi_2} s_{\vartheta_2}, 2c^2_{\vartheta_2} c_{\varphi_2} 
s_{\varphi_2}]^\top}$, $(\mathbf{p}_2 - \mathbf{p}_1)/ \lVert \mathbf{p}_2 - \mathbf{p}_1 \rVert = 
[d_{2,1}^{(1)}, d_{2,1}^{(2)}, d_{2,1}^{(3)}]^\top$, and $\splitatcommas{\mathbf{v}_{2,1} = [1, 
(d_{2,1}^{(1)})^2, (d_{2,1}^{(2)})^2, (d_{2,1}^{(3)})^2, d_{2,1}^{(1)} d_{2,1}^{(2)}, d_{2,1}^{(1)} 
d_{2,1}^{(3)},  d_{2,1}^{(2)} d_{2,1}^{(3)} ]^\top}$. In a similar manner:
\begin{eqnarray} \label{eq:prob.10}
	\begin{split}
		G_1^2(\vartheta_{2,1}^A) & \approx D_d^2 \, \mathbf{g}(\vartheta_1, \varphi_1)^\top 
		\mathbf{v}_{2,1}, \\
		G_1^2(\vartheta_{1,0}^D) & \approx D_d^2 \, \mathbf{g} (\vartheta_1, \varphi_1)^\top 
		\mathbf{v}_{1,0}. \label{eq:prob.11}
	\end{split}
\end{eqnarray}
Note that the antenna gains~\eqref{eq:prob.10} depend not only on the position of the~\ac{UAV}-$1$ 
(i.e., $\mathbf{p}_1$) but also on its roll ($\varphi_1$) and pitch ($\vartheta_1$) angles. Now, we 
are optimizing the~\ac{UAV}-$1$ trajectory w.r.t. its position ($\mathbf{p}_1$), velocity 
($\mathbf{v}_1$) and acceleration ($\mathbf{a}_1)$; these variables implicitly determine the roll 
and pitch angles throughout the full~\ac{UAV}-$1$ trajectory~\cite{Siciliano2016Handbook, 
Mueller2015TRO}. Nevertheless, considering such nonlinear relation would significantly complicate 
the optimization problem and thus for simplicity, during the optimization, we will 
evaluate~\eqref{eq:prob.10} at $\vartheta_1=\varphi_1=0$ which corresponds to the hovering position.

Thus, after all the elements discussed in this section we reformulate the optimization 
problem~\eqref{eq:optimizationProblem} as:
\begin{equation} \label{eq:optimizationProblemNew}
	\begin{split}
		&\minimize_{\mathbf{p}_1, \mathbf{v}_1, \mathbf{a}_1}\ \ 
		\displaystyle\sum_{k=0}^N \Biggl({ \frac{1}{\log_2^p \Bigl(1 + 
		\prescript{k}{}{\xi}_{1,0} \Bigr)}}, { \frac{1}{\log_2^p \Bigl(1 + 
		\prescript{k}{}{\xi}_{2,1} \Bigr)}} \Biggr)^{1/p} \\
		&\quad \,\; \text{s.t.}~\quad \prescript{k}{}{\xi}_{1,0} = \frac{D_B^2 \, D^2 
		\mathbf{g}(\vartheta_1, \varphi_1)^\top \mathbf{v}_{1,0} P}{\lVert 
		\prescript{k}{}{\mathbf{p}}_1 - \mathbf{p}_0 \rVert ^2\sigma^2_0},\\
		&\qquad\qquad \prescript{k}{}{\xi}_{2,1} = \frac{D^4 \mathbf{g}(\vartheta_1, 
		\varphi_1)^\top \mathbf{v}_{2,1} \mathbf{g}(\vartheta_2, \varphi_2)^\top \mathbf{v}_{2,1} 
		P} {\lVert{ \prescript{k}{}{\mathbf{p}}_2 - \prescript{k}{}{\mathbf{p}}_1 
		\rVert^2 \, \sigma^2_1}},\\
		&\qquad \;\;\,~\quad\; \lvert \prescript{k}{}{\mathbf{v}}^{(j)} \rvert \leq 
		\mathbf{v}^{(j)}_\mathrm{max}, \lvert \prescript{k}{}{\mathbf{a}}^{(j)} \vert  \leq 
		\mathbf{a}^{(j)}_\mathrm{max}, \\
		&\,\, \qquad \qquad \text{eq.}~\eqref{eq:splines}, \forall k=\{0,1, \dots, N-1\}, \\
		&\,\, \qquad \qquad \text{with} \ \vartheta_1, \varphi_1 = 0.
	\end{split}
\end{equation}



\vspace{-1em}
\section{Simulations}
\label{sec:Sim}

In this section, we test the trajectory of~\ac{UAV}-$1$ optimized using the proposed method. The 
trajectory of~\ac{UAV}-$2$, $\mathcal{T}_2$, is taken from~\cite{Silano2021ICRARAL}; it was 
designed for a power tower inspection task, and was experimentally validated. The position of the 
BS is $[ 1.00,  3.00,  1.50]^\top$ and the initial positions for~\ac{UAV}-$1$ and~\ac{UAV}-$2$ are 
$[ 0.00,  3.00,  1.50]^\top$ and $[ 4.00,  3.00,  1.50]^\top$, respectively.

The optimization problem \eqref{eq:optimizationProblemNew} is solved using the CasADi 
library\footnote{\url{https://web.casadi.org}} and NLP\footnote{\url{http://cvxr.com}} as solver. 
The maximum number of iterations is set to $2000$, with acceptable tolerance of~$\num{1.0e-4}$. 
Simulations were carried out using the 2019b release of MATLAB on a laptop with an i7-8565U 
processor (1.80 GHz) and 32GB of RAM running on Ubuntu 18.04. Videos with the experiments and 
numerical simulations in MATLAB and Gazebo are available 
at~\url{http://mrs.felk.cvut.cz/optimum-trajectory-relay}.

\begin{figure}
	\begin{center}
		\vspace*{-8mm}
		\adjincludegraphics[trim={{.075\width} {.0\height} {0.0\width}			
			{.0\height}},clip,scale=0.205]{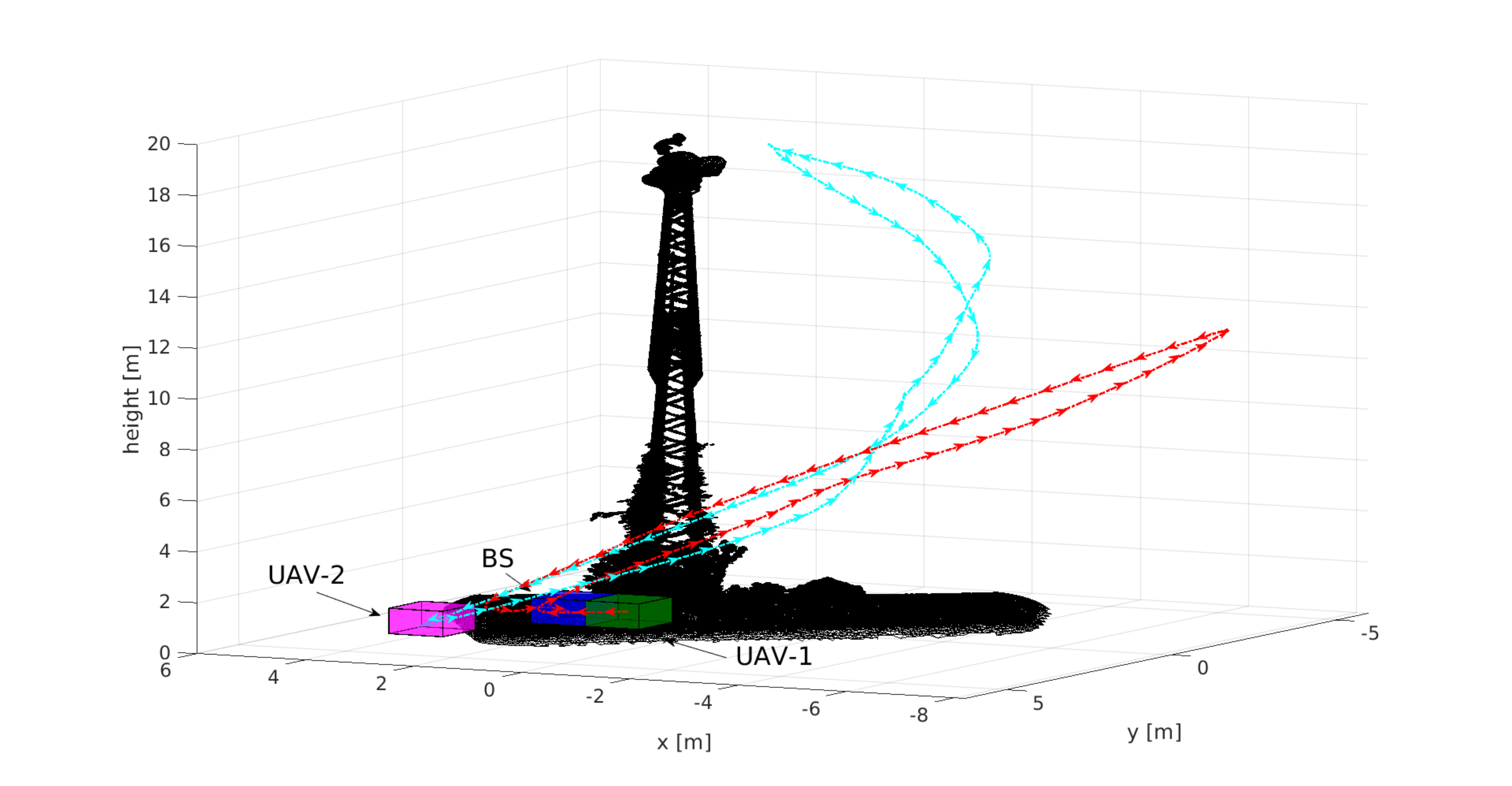}
		\vspace*{-9mm}
	\end{center}
	\caption{First~\ac{UAV}-$1$ trajectory (red) and~\ac{UAV}-$2$ trajectory (cyan). 
	The~\ac{BS},~\ac{UAV}-$1$ and~\ac{UAV}-$2$ starting points are reported in blue, green and 
	magenta boxes, respectively.}
	\label{fig:scenario253D}
	\vspace{-1em}
\end{figure}

\begin{figure}
	\begin{center}
		\vspace*{-2mm}
		\adjincludegraphics[trim={{.075\width} {.0\height} {0.0\width}			
			{.0\height}},clip,scale=0.205]{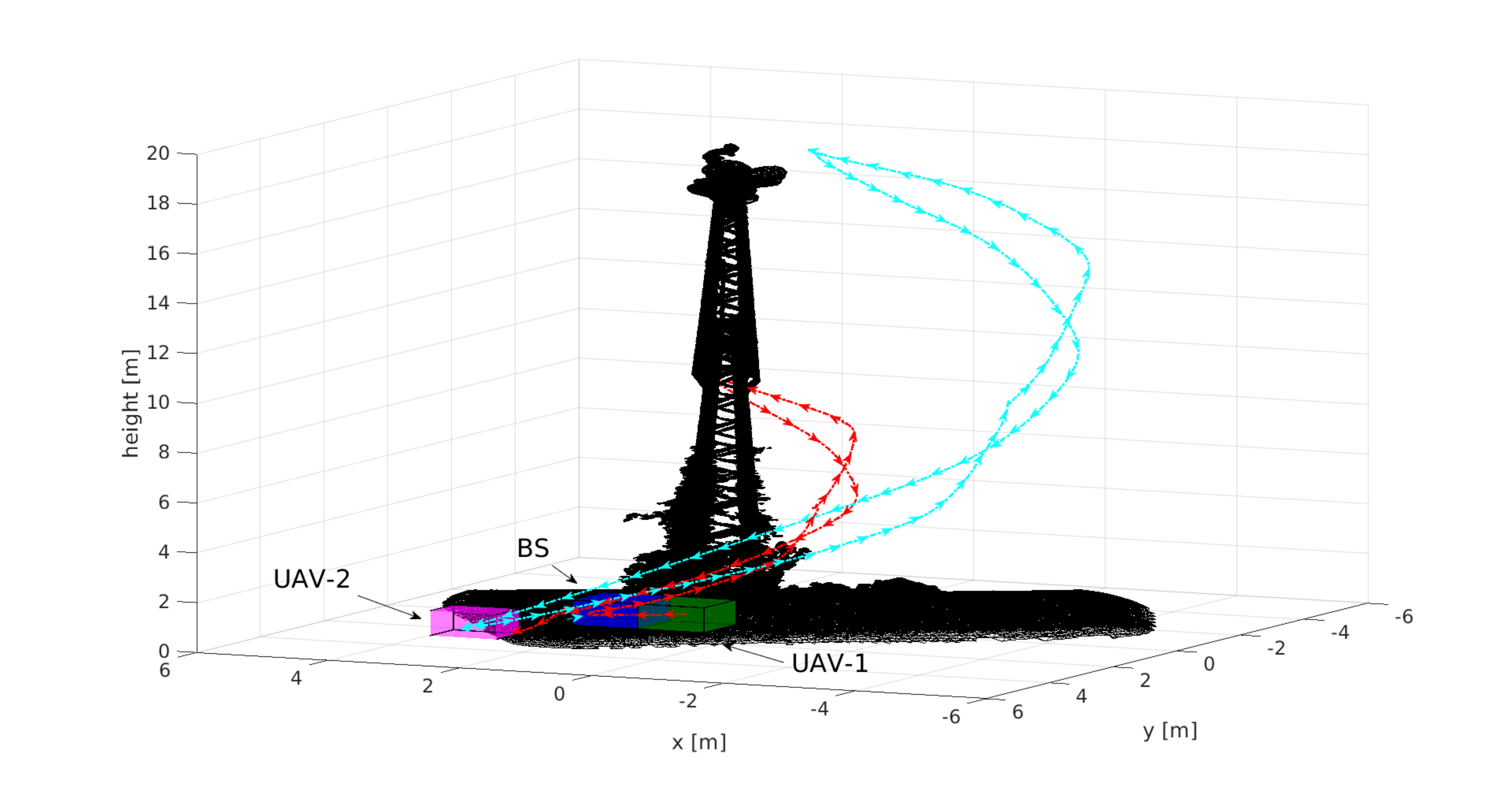}
		\vspace*{-9mm}
	\end{center}
	\caption{Second~\ac{UAV}-$1$ trajectory (red) and~\ac{UAV}-$2$ trajectory (cyan).}
	\label{fig:scenario263D}
	\vspace{-1em}
\end{figure}

\begin{figure}
	\begin{center}
		\scalebox{0.92}{
			\begin{tikzpicture}
			\begin{axis}[%
			width=2.8119in,%
			height=1.4183in,%
			at={(0.758in,0.481in)},%
			scale only axis,%
			xmin=0,%
			xmax=18,%
			ymax=30,%
			ymin=-30,%
			xmajorgrids,%
			ymajorgrids,%
			ylabel style={yshift=-0.415cm}, 
			xlabel={Time [\si{\second}]},%
			ylabel={$\varphi$ [\si{\deg}]},%
			axis background/.style={fill=white},%
			legend style={at={(0.55,1.05)},anchor=north,legend cell align=left,draw=none,legend 
			columns=-1,align=left,draw=white!15!black}
			]
			\addplot [color=blue, dashed, line width=0.75pt] 
			file{matlabPlots/trajectory3/phi_UAV2.txt};%
			\addplot [color=red, dotted, line width=0.75pt] 
			file{matlabPlots/trajectory3/phi_UAV1.txt};%
			\addplot [color=green, dashdotted, line width=0.75pt] 
			file{matlabPlots/trajectory4/phi_UAV1.txt};%
			\legend{\ac{UAV}-$2$, \ac{UAV}-$1_\mathrm{tr1}$, \ac{UAV}-$1_\mathrm{tr2}$};%
			\end{axis}
			\end{tikzpicture}
		}
	\end{center}
	\vspace*{-4mm}
	\caption{Roll angle ($\varphi$) for the~\ac{UAV}-$2$ and both~\ac{UAV}-$1$ trajectories (i.e., 
	$\mathrm{UAV}\text{-}1_\mathrm{tr1}$ and $\mathrm{UAV}\text{-}2_\mathrm{tr2}$).}
	\label{fig:rollangles}
\end{figure}
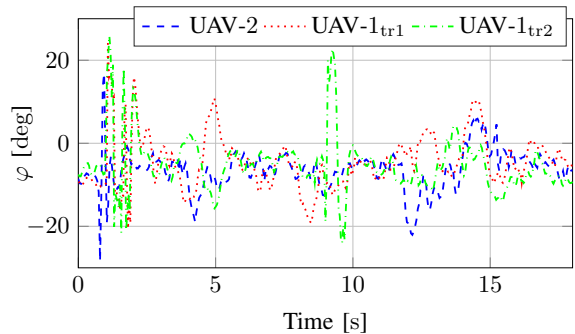

\begin{figure}
	\begin{center}
		\scalebox{0.92}{
			\begin{tikzpicture}
			\begin{axis}[%
			width=2.8119in,%
			height=1.4183in,%
			at={(0.758in,0.481in)},%
			scale only axis,%
			xmin=0,%
			xmax=18,%
			ymax=30,%
			ymin=-30,%
			xmajorgrids,%
			ymajorgrids,%
			ylabel style={yshift=-0.415cm}, 
			xlabel={Time [\si{\second}]},%
			ylabel={$\vartheta$ [\si{\deg}]},%
			axis background/.style={fill=white},%
			legend style={at={(0.55,0.15)},anchor=north,legend cell align=left,draw=none,legend 
			columns=-1,align=left,draw=white!15!black}
			]
			\addplot [color=blue, dashed, line width=0.75pt] 
			file{matlabPlots/trajectory4/theta_UAV2.txt};%
			\addplot [color=red, dotted, line width=0.75pt] 
			file{matlabPlots/trajectory3/theta_UAV1.txt};%
			\addplot [color=green, dashdotted, line width=0.75pt] 
			file{matlabPlots/trajectory4/theta_UAV1.txt};%
			\legend{\ac{UAV}-$2$, \ac{UAV}-$1_\mathrm{tr1}$, \ac{UAV}-$1_\mathrm{tr2}$};%
			\end{axis}
			\end{tikzpicture}
		}
	\end{center}
	\vspace*{-4mm}
	\caption{Pitch angle ($\vartheta$) for the~\ac{UAV}-$2$ and both~\ac{UAV}-$1$ trajectories.}
	\label{fig:pitchangles}
\end{figure}
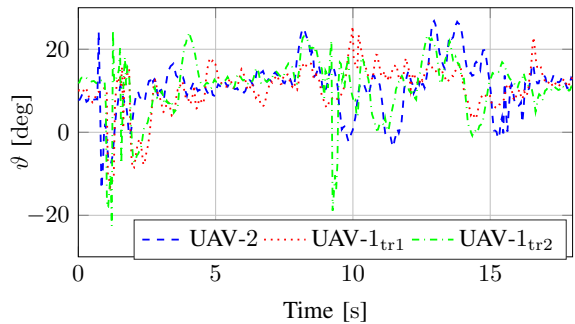

To show the advantages of considering the antenna orientation changes, due to the~\ac{UAV} tilt in 
the trajectory optimization we compare two different trajectories: the first trajectory is obtained 
after solving~\eqref{eq:optimizationProblemNew}, see Fig.~\ref{fig:scenario253D}; the second 
trajectory is obtained after solving \eqref{eq:optimizationProblemNew} but disregarding the antenna 
radiation patterns\footnote{This is done by exchanging $\mathbf{g}(\vartheta_j, \varphi_j)^\top 
\mathbf{v}_{k,j}$ with $1$.}, see Fig.~\ref{fig:scenario263D}. The time taken by the solver for the 
first and second trajectories are \SI{154}{\second} and~\SI{63}{\second}, respectively. Both 
trajectories are optimized using the same~\ac{UAV}-$2$ trajectory $\mathcal{T}_2$ and the following 
parameter values: $p=10$, $\nicefrac{D^4P}{\sigma^2_1}=10^9$,  
$\nicefrac{D_B^2D^2P}{\sigma^2_0}=10^9$, $\mathbf{v}_\mathrm{max}=\SI{2}{\meter\per\second} 
$, $\mathbf{a}_\mathrm{max}$ = $\SI{2}{\meter\per\square\second}$, sampling time 
$T_s=\SI{0.05}{\second}$ and trajectory duration $T=\SI{18.0}{\second}$, and $\alpha$, $\beta$, and 
$\gamma$ from~\cite[eq.~(63)]{Mueller2015TRO}. After obtaining the trajectories in MATLAB we 
execute them in Gazebo\footnote{Gazebo is a robotics simulator that uses complex and realistic 
models for the~\acp{UAV}; we exploit the advantages of~\ac{SIL} simulations~\cite{Silano2019SMC} by 
generating the trajectories in MATLAB and then testing them in Gazebo.} to verify their feasibility 
and to measure the corresponding realistic roll and pitch angles for both~\ac{UAV}-$1$ trajectories 
and for~\ac{UAV}-$2$ trajectory $\mathcal{T}_2$, see Figs.~\ref{fig:rollangles} 
and~\ref{fig:pitchangles}. These angles provide the reader with an idea about the~\acp{UAV} antenna 
orientation. 

From Figs.~\ref{fig:scenario253D} and~\ref{fig:scenario263D} we note that the first~\ac{UAV}-$1$ 
trajectory seems to get away from the~\ac{UAV}-$2$ and the~\ac{BS} to align itself with the 
the~\ac{UAV}-$2$'s antenna; the second~\ac{UAV}-$1$ trajectory simply mimics $\mathcal{T}_2$ in a 
scaled manner.

\begin{figure}
	\begin{center}
		\hspace{-0.725cm}
		\begin{subfigure}{0.45\columnwidth}
			\scalebox{0.52}{
				\begin{tikzpicture}
				\begin{axis}[%
				width=2.8119in,%
				height=1.8183in,%
				at={(0.758in,0.481in)},%
				scale only axis,%
				xmin=0,%
				xmax=18,%
				ymax=40,%
				ymin=20,%
				xmajorgrids,%
				ymajorgrids,%
				ylabel style={yshift=-0.415cm}, 
				xlabel={Time [\si{\second}]},%
				ylabel={Normalized Bit rate},%
				axis background/.style={fill=white},%
				legend style={at={(0.5,0.95)},anchor=north,legend cell align=left,draw=none,legend 
				columns=-1,align=left,draw=white!15!black}
				]
				\addplot [color=blue, solid, line width=0.75pt] 
				file{matlabPlots/trajectory3/SNR_UAV1.txt};%
				\addplot [color=red, solid, line width=0.75pt] 
				file{matlabPlots/trajectory3/SNR_UAV2.txt};%
				\legend{$\mathrm{UAV1}\rightarrow\mathrm{BS}$, $\mathrm{UAV2}\rightarrow 
				\mathrm{UAV1}$};%
				\node [draw,fill=white] at (rel axis cs: 0.925,0.1) {\shortstack[l]{tr1}};
				\end{axis}
				\end{tikzpicture}
			}
		\end{subfigure}
		\hspace{0.25cm}
		\begin{subfigure}{0.45\columnwidth}
			\scalebox{0.52}{
				\begin{tikzpicture}
				\begin{axis}[%
				width=2.8119in,%
				height=1.8183in,%
				at={(0.758in,0.481in)},%
				scale only axis,%
				xmin=0,%
				xmax=18,%
				ymax=40,%
				ymin=20,%
				xmajorgrids,%
				ymajorgrids,%
				ylabel style={yshift=-0.415cm}, 
				xlabel={Time [\si{\second}]},%
				ylabel={Normalized Bit rate},%
				axis background/.style={fill=white},%
				legend style={at={(0.5,0.95)},anchor=north,legend cell align=left,draw=none,legend 
				columns=-1,align=left,draw=white!15!black}
				]
				\addplot [color=blue, solid, line width=0.75pt] 
				file{matlabPlots/trajectory4/SNR-UAV1.txt};%
				\addplot [color=red, solid, line width=0.75pt] 
				file{matlabPlots/trajectory4/SNR-UAV2.txt};%
				\legend{$\mathrm{UAV1}\rightarrow\mathrm{BS}$, $\mathrm{UAV2}\rightarrow 
				\mathrm{UAV1}$};%
				\node [draw,fill=white] at (rel axis cs: 0.925,0.1) {\shortstack[l]{tr2}};
				\end{axis}
				\end{tikzpicture}
			}
		\end{subfigure}
	\end{center}
	\vspace*{-4mm}
	\caption{Instantaneous bit rates of both links for both trajectories: first~\ac{UAV}-$1$ 
	trajectory (left), second~\ac{UAV}-$1$ trajectory (right).}
	\label{fig:SNR}
\end{figure}
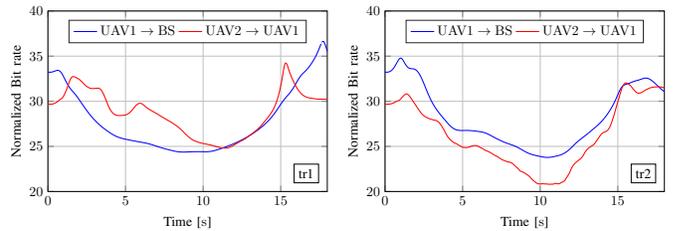

The number of bits that can be transmitted, calculated with the optimization target of 
\eqref{eq:optimizationProblemNew}, using the first~\ac{UAV}-$1$ trajectory is $6.4$\% higher than 
that for the second~\ac{UAV}-$1$ trajectory. In Fig.~\ref{fig:SNR} we see the instantaneous 
normalized bit rate of both links for both trajectories: the minimum instantaneous bit rate for the 
first~\ac{UAV}-$1$ trajectory is $17.3$\% higher than that for the second~\ac{UAV}-1 trajectory. 

Higher gain antennas have lower half-power bandwidth; as the antenna gain increases the~\ac{UAV} 
relay performance becomes more sensitive to the antenna orientation changes. Additionally, as the 
multi-rotor~\ac{UAV} moves faster its tilt (and that of its antenna) increases. Hence, higher gain 
antennas and/or higher~\ac{UAV} speeds increase the relevance of our proposed technique.



\section{Conclusions}
\label{sec:Conc}

We presented a framework to design a multi-rotor trajectory considering a realistic dynamic model 
of the aircraft and the antenna orientation changes due to the~\ac{UAV} tilt. Initial results show 
that considering the antenna orientation changes during the trajectory optimization can increase 
the overall number of bits transmitted and the minimum instantaneous bit rate. The actual benefit 
derived from this consideration depends on the type of antenna and on the~\ac{UAV} trajectories. 
Future research will determine the characteristics of the trajectories that benefit the most from 
this approach and also will evaluate the energy efficiency of the resulting trajectories. 



\begin{appendices}
	
	\section{Smooth approximation for $\min$ function}
	\label{Appendix:minApp}
	
	In this appendix we present a smooth approximation for $\max_{j}(\varepsilon_j)$. From the 
	definition of the $H_{\infty}$ norm we have:
	\begin{equation} \label{eq:ap1_eq1}
		\lVert \varepsilon_1, \varepsilon_2, \dots, \varepsilon_N \rVert_\infty = \max_{j \in \{1, 
		2, \dots, N\}} \varepsilon_j.
	\end{equation}
	Now, it is well known that $p\text{-norm} \rightarrow H_\infty$ norm as $p\rightarrow 
	\infty$~\cite{Stewart1973Book}. In other words, $\lVert \mathbf{u} \rVert_{\infty} = {\Lim{p 
	\rightarrow \infty} \lVert \mathbf{u} \rVert_p}$, and thus:
	\begin{equation} \label{eq:ap1_eq2}
		\resizebox{0.89\hsize}{!}{$
			\lVert \varepsilon_1, \varepsilon_2, \dots, \varepsilon_N \rVert_\infty \approx \lVert 
			\varepsilon_1, \varepsilon_2, \dots, \varepsilon_N \rVert_p = \left( \sum_{j = 1}^N 
			\varepsilon_j^p \right)^{\frac{1}{p}},
			$}
	\end{equation}
	where the r.h.s. of~\eqref{eq:ap1_eq2} gets closer to the l.h.s. as $p \in \mathbb{N}^+$ 
	increases. Finally, we have $\max_{j \in \{1, 2, \dots, N\}} \varepsilon_j \approx \left( 
	\sum_{j=1}^N \varepsilon_j^{p} \right)^{\frac{1}{p}}.$
	
	
	
	\section{Antenna Radiation Pattern approximation}
	\label{Appendix:radApp}
	
	\begin{figure}
		\begin{center}
			\scalebox{0.9}{
				\begin{tikzpicture}
				\begin{axis}[%
				width=2.8119in,%
				height=1.1183in,%
				at={(0.758in,0.481in)},%
				scale only axis,%
				xmin=0,%
				xmax=6.28,
				ymin=0,%
				ymax=2.75,%
				xmajorgrids,%
				ymajorgrids,%
				ylabel style={yshift=-0.415cm}, 
				xlabel={$\vartheta$ [\si{\radian}]},%
				ylabel={Antenna Power Gain},%
				axis background/.style={fill=white},%
				legend style={ nodes={scale=0.7, transform shape}, at={(0.25,0.975)}, anchor=north, 
				align=left, draw=none, draw=white!15!black }
				]
				\addplot [color=blue, solid, line width=0.75pt] 
				file{matlabPlots/radiationPattern/eta.txt};%
				\addplot [color=blue, dashed, line width=0.75pt] 
				file{matlabPlots/radiationPattern/eta_approximated.txt};%
				\legend{$G^2(\vartheta + \pi/2)$, $\cos^2(\vartheta)$};%
				\end{axis}
				\end{tikzpicture}
			}
		\end{center}
		\vspace*{-4mm}
		\caption{Half-wave dipole power radiation pattern and its approximation.}
		\label{fig:antennaApproximation}
	\end{figure}
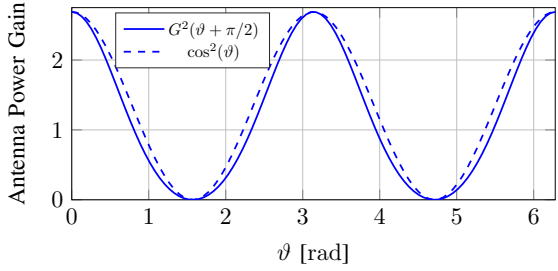
	
	Let us consider the~\ac{UAV}-$2\rightarrow$\ac{UAV}-$1$ channel and define the vector 
	$\mathbf{d}_{2,1}\triangleq(\mathbf{p}_2 - \mathbf{p}_1)/ \lVert \mathbf{p}_2 - \mathbf{p}_1 
	\rVert =[d_{2,1}^{(1)}, d_{2,1}^{(2)}, d_{2,1}^{(3)}]^\top$, where $\mathbf{d}_{2,1}$ is 
	expressed within the global coordinate system $O_W$. Then let us define 
	$\underline{\mathbf{d}}_{2,1}\triangleq \prescript{W}{}{\mathbf{R}}_{B_q} \mathbf{d}_{2,1}$ 
	with the rotation matrix $\prescript{W}{}{\mathbf{R}}_{B_q}$ given as 
	in~\cite{Siciliano2016Handbook}. Since $\lVert \mathbf{d}_{2,1} \rVert = 1$ and the rotation 
	matrix is unitary then $\lVert \underline{\mathbf{d}}_{2,1} \rVert = 1$. Let 
	$\hat{\underline{\mathbf{d}}}_{2,1}$ be the projection of $\underline{\mathbf{d}}_{2,1}$ on the 
	plane $\mathcal{H}_q$ spanned by $O_{B_q}{\xx}_{B_q}$ and $O_{B_q}{\yy}_{B_q}$. Now, since we 
	are working on an Euclidean space~\cite{Gantmacher1977Book} the inner product between 
	$\hat{\underline{\mathbf{d}}}_{2,1}$ and $\underline{\mathbf{d}}_{2,1}$ is:
	\begin{equation} \label{app2:eq:1}
		{\langle \hat{\underline{\mathbf{d}}}_{2,1}, \underline{\mathbf{d}}_{2,1} \rangle }/{ 
		\lVert \hat{\underline{\mathbf{d}}}_{2,1} \rVert} = \cos(\varphi_{2,1}),
	\end{equation}
	where $\varphi_{2,1}$ is the angle formed between $\hat{\underline{\mathbf{d}}}_{2,1}$ and 
	$\underline{\mathbf{d}}_{2,1}$ that corresponds to the~\ac{AoD} measured w.r.t. 
	$\mathcal{H}_q$, and $\vartheta_{2,1}$ in~\eqref{CS:2} is the~\ac{AoD} measured w.r.t. 
	$O_{B_q}{\zz}_{B_q}$. Thus, we have $\cos(\varphi_{2,1}) = \cos(\varphi_{2,1} + \pi/2)$. 
	$\langle \hat{\underline{\mathbf{d}}}_{2,1}, \underline{\mathbf{d}}_{2,1} \rangle = 
	\hat{\underline{\mathbf{d}}}_{2,1}^\top {\underline{\mathbf{d}}}_{2,1}$ is the inner product of 
	both vectors in the Euclidean space. After some algebra:
	\begin{equation} \label{app2:eq:2}
		{\langle \hat{\underline{\mathbf{d}}}_{2,1}, {\underline{\mathbf{d}}}_{2,1} \rangle} = 
		\lVert \hat{\underline{\mathbf{d}}}_{2,1} \rVert^2 = \mathbf{g}(\vartheta_2, 
		\varphi_2)^\top \mathbf{v}_{2,1},
	\end{equation}
	where $\splitatcommas{\mathbf{g}(\vartheta_2, \varphi_2) = [1, -s^2_{\vartheta_2}, 
	-c^2_{\vartheta_2} s^2_{\varphi_2}, -c^2_{\varphi_2} c^2_{\vartheta_2}, 2c_{\vartheta_2} 
	s_{\vartheta_2} s_{\varphi_2}, -2c_{\vartheta_2} c_{\varphi_2} s_{\vartheta_2}, 
	2c^2_{\vartheta_2} c_{\varphi_2} s_{\varphi_2}]^\top}$ and $\splitatcommas{\mathbf{v}_{2,1} = 
	[1, (d_{2,1}^{(1)})^2, (d_{2,1}^{(2)})^2, (d_{2,1}^{(3)})^2, d_{2,1}^{(1)} d_{2,1}^{(2)}, 
	d_{2,1}^{(1)} d_{2,1}^{(3)},  d_{2,1}^{(2)} d_{2,1}^{(3)} ]^\top}$. It can be observed from 
	Fig.~\ref{fig:antennaApproximation} that $G_2^2(\vartheta_{2,1}+\pi/2) \approx 
	D^2\cos^2(\vartheta_{2,1})$; then, considering (\ref{app2:eq:1}) and (\ref{app2:eq:2}) we 
	obtain $G_2^2(\vartheta_{2,1} + \pi/2) \approx D^2 \mathbf{g}(\vartheta_2, \varphi_2)^\top 
	\mathbf{v}_{2,1}$.
	
\end{appendices}



\vspace{-0.4em}
\bibliographystyle{IEEEtran}
\bibliography{bib.bib}

\end{document}